\documentclass[11pt]{article}

\usepackage[utf8]{inputenc}
\usepackage[T1]{fontenc}
\usepackage{lmodern}
\usepackage[margin=1in]{geometry}
\usepackage{graphicx}
\usepackage{booktabs}
\usepackage{amsmath}
\usepackage{amssymb}
\usepackage{hyperref}
\usepackage{xcolor}
\usepackage{caption}
\usepackage{subcaption}
\usepackage{natbib}
\usepackage{authblk}
\usepackage{titlesec}

\hypersetup{
    colorlinks=true,
    linkcolor=blue!60!black,
    citecolor=blue!50!black,
    urlcolor=blue!60!black
}

\newcommand{\fig}[2]{\IfFileExists{#2}{\includegraphics[width=#1]{#2}}{\fbox{\parbox{0.8\textwidth}{\centering\color{red}missing figure: \texttt{\detokenize{#2}}}}}}

\title{\textbf{Structural Pattern Mining in Inka Khipus: An Audit of Clustering and Provenance Classification in the Open Khipu Repository}\thanks{Version 2 (November 2026). This version withdraws the typological interpretation of the clusters and of the imperial-style classification, corrects the reported label counts, and adds sequential-exclusion, institution-grouped and missingness analyses.}}

\author{Maria Contreras}
\affil{Independent Researcher (Disentangle Lab)}
\date{\today}

\begin{document}

\maketitle

\begin{abstract}
\noindent
Khipus---knotted cord devices---were the primary recording medium of the Inka
Empire (c.\ 1400--1532 CE), yet their system remains undeciphered. We audit what
unsupervised and supervised learning can say about them, using the Open Khipu
Repository (OKR v2.1.0; 619 khipus, 27 structural features). UMAP and HDBSCAN
yield three clusters (17/159/443; silhouette 0.82 in the embedding but 0.06 in
feature space), and a provenance classifier reaches a weighted $F_1$ of 0.46
under stratified cross-validation. Both track how khipus were recorded rather
than how they were made. Cord-level twist coding splits the clusters into three
regimes (blank, unspecified/Z, S), and a depth-two tree on documentation
indicators predicts cluster membership (accuracy 0.914 against a 0.716 majority
baseline). In feature space, a permutation null is not rejected on the full
corpus ($p=0.949$), after removing empty records ($n=612$, $p=0.823$), or on a
clean subset ($n=430$, $p=0.926$). The only rejection occurs in an intermediate
two-cluster subset ($n=467$, $p=0.009$; not significant in UMAP space,
$p=0.103$), and it disappears when 37 records uploaded under one OKR
contributor account are removed. All 18 Inka Late Horizon labels belong to one
museum and one cataloger prefix; 14 of them are empty or knotless records, and
a classifier on structural features recovers exactly those 14
($F_1=0.82$), so the label cannot be evaluated as an imperial signature. A
recount of the recto/verso pattern of six Santa Valley khipus from the same
transcriptions reproduces the pattern reported by \citet{medrano2018}, but is
not independent evidence. We release exclusion lists, null-model tables and
code, and argue that structural claims about khipus need explicit controls for
missingness and record provenance.

\vspace{0.5em}
\end{abstract}

% ─────────────────────────────────────────────────────────────────────────────
\section{Introduction}
\label{sec:intro}

The Inka Empire was the largest polity in the pre-Columbian Americas,
administering millions of people across the Andes without a writing system in
the conventional sense. In its place, the Inka state relied on \emph{khipus}
(Quechua for ``knot''): assemblages of spun and plied cotton or camelid-fiber
cords, knotted to record numerical and possibly narrative information. Spanish
chroniclers described specialist record-keepers, the \emph{khipukamayuqs}, who
read these devices as administrative documents.

Roughly 1{,}000 khipus survive in museum and private collections worldwide.
While the base-ten numerical convention of many khipus is well understood
\citep{ascher1997, urton2003}, whether and how they encode language, names, or
narratives remains one of the principal open problems in Andean studies. Most
quantitative work to date has focused either on individual specimens or on
matching numerical sums across related khipus \citep{urton2005}. The
\emph{structural} dimensionality of the full corpus---what construction
conventions distinguish khipus across regions and periods---has received less
systematic computational attention.

This paper applies machine-learning methods to the Open Khipu Repository (OKR),
the most comprehensive public khipu database, and tests whether the structure
they find survives the recording and cataloging history of the corpus. Our
contributions are:

\begin{enumerate}
    \item A reproducible feature pipeline that turns the OKR's relational
    structure into a 27-dimensional structural vector per khipu, together with
    separate documentation-completeness indicators
    (Section~\ref{sec:features}).
    \item A sequential-exclusion audit ($619\to612\to467\to430$) showing that
    the three clusters do not survive removal of empty, poorly documented and
    single-contributor records, with null models in feature space
    (Section~\ref{sec:clustering}).
    \item Evidence that cluster membership is predictable from documentation
    features, and that the Late Horizon label coincides with one museum, one
    cataloger prefix and a set of empty or knotless records
    (Section~\ref{sec:classification}).
    \item A recount of the Santa Valley recto/verso pattern from the public OKR,
    which reproduces \citet{medrano2018} using the same transcriptions
    (Section~\ref{sec:santavalley}).
\end{enumerate}

% ─────────────────────────────────────────────────────────────────────────────
\section{Related Work}
\label{sec:related}

The systematic study of khipus as data begins with \citet{ascher1997}, whose
descriptive code underlies modern databases. The Harvard Khipu Database Project,
initiated by Urton and Brezine, formalized the recording of knot, cord, and
color attributes at scale; \citet{urton2005} used it to demonstrate hierarchical
accounting relationships among the Puruchuco khipus, where lower-level cord sums
reappear on higher-level khipus. \citet{urton2015} described the Inkawasi
khipus, archaeologically associated with stored agricultural goods.

On the decipherment front, \citet{medrano2018} matched six khipus from the
Santa Valley to a 1670 colonial tribute census, arguing that first-cord knot
sums correspond to assessed tribute and that cord attachment direction
(recto/verso) encodes moiety (\emph{hanan}/\emph{hurin}) affiliation.
\citet{fitzpatrick2024} subsequently refined this alignment.
\citet{hyland2017} proposed a phonetic reading of the Collata khipus using
community knowledge. On the computational side, \citet{clindaniel2019} applied
network and statistical methods to OKR structure, and \citet{medrano2021}
built a corpus-linguistic typology of colonial khipu transcriptions.

The naming prefixes of khipus (e.g., UR, AS, HP, QU) identify the researcher
who cataloged each specimen rather than its site or period
\citep{brezine2023}. In a companion re-analysis of the Khipu Field Guide
database, we showed that cataloger identity and documentation completeness are
confounds that can generate apparent structure which disappears under
null-model testing \citep{contreras2026kfg}. The present paper applies the same
scrutiny to the OKR, and revises our own earlier reading of it.

% ─────────────────────────────────────────────────────────────────────────────
\section{Dataset and Feature Engineering}
\label{sec:features}

\subsection{The Open Khipu Repository}

The OKR is distributed as a serverless SQLite database containing data on
619 khipus, 54{,}403 cords, and 110{,}677 knots,
aggregated from the work of Ascher, Pereyra, the Harvard Khipu Database Project,
and others. We query four principal tables: \texttt{khipu\_main} (specimen
metadata: provenance, region, museum, catalog name, uploading account),
\texttt{cord} (per-cord length, twist, hierarchical level, attachment),
\texttt{knot} (per-knot type, direction, turns), and \texttt{ascher\_cord\_color}
(color encoding).

\paragraph{Records without data.}
Seven khipus (KH0601--KH0607; Dallas Museum of Art) have no cord, knot or color
records. A further seven Dallas khipus (KH0593--KH0599) have 18--160 cords but
no knots and no colors; the OKR team indicated that this reflects a data-entry
issue under correction (personal communication, November 2026). The first group
is excluded in the $n=612$ analyses; the second remains in all analyses and
falls in cluster~0.

\paragraph{Region labels.}
A region value is recorded for most khipus, but most of these are empty strings
(388) or ``unknown'' (14): only 120 of the 619 khipus (113 of 612) carry an
informative region label. An earlier version of this paper reported 522 labeled
specimens and 135 used for classification; the first count included empty
strings that were treated as labels in memory but not after a file round-trip,
and both counts are superseded by the figures given here.

\paragraph{Sequential exclusions.}
We apply three filters in order and export the excluded records with their
reason (\texttt{results/excluded\_khipus\_okr\_2.1.0.csv}): (i) empty records
with zero cords ($n=7$, $619\to612$); (ii) khipus with at least $20\%$ of cords
of unrecorded twist ($n=145$, $\to467$); and (iii) records uploaded under one
OKR contributor account (the \texttt{CREATED\_BY} field, $n=37$, $\to430$; the
account is identified in the released exclusion list). In the two sites
Incahuasi and Armatambo/Huaca San Pedro, twist coding in records from that
account contrasts with that of other records from the same sites
(\texttt{results/table\_b7\_jclindaniel\_sites.csv}). The cause is unresolved,
and we make no claim of error by any contributor: we treat the twist of those
records as non-comparable, and we never recode or invert S and Z in any subset.
The $20\%$ threshold is arbitrary, and Section~\ref{sec:thresh} varies it.

\subsection{Features}

We aggregate cord- and knot-level attributes into a 27-dimensional structural
vector per khipu, rebuilt from the public OKR tables by the released notebook
and identical to the feature set of the first version of this paper:

\begin{itemize}
    \item \textbf{Cord structure:} number of cords, mean and standard deviation
    of cord length, maximum cord level, number and ratio of subsidiary cords.
    \item \textbf{Twist:} counts of S- and Z-direction cords and their ratio.
    \item \textbf{Knots:} total knot count, mean turns, knot-direction counts
    and ratio, and per-type knot counts.
    \item \textbf{Color:} number of unique colors, color entropy, and a
    multicolor indicator.
\end{itemize}

Color entropy $H=-\sum_i p_i\log p_i$ captures palette diversity, where $p_i$ is
the relative frequency of color $i$ on a given khipu. Missing values are imputed
as zero, as in the first version, and features are standardized before
unsupervised analysis. A reconstructed 23-feature variant is also released
(mode \texttt{robustez} of the notebook, \texttt{results/table\_sensitivity\_matrix.csv}).

\paragraph{Documentation indicators.}
Because twist, knot direction and attachment side are frequently unrecorded,
and zero-imputation turns unrecorded values into zeros, the structural features
are partly proxies for how completely a khipu was recorded. We therefore compute
separate indicators---the share of cords with unrecorded twist, the share of
knots with unrecorded direction, and the share of cords with unrecorded
attachment side---and keep them out of the structural feature set. An extended
set of eight indicators splits unrecorded twist into explicit ``U'' codes and
blank/null entries, and adds the share of cords without recorded length, color
rows per cord, and knot rows per cord. They are used only to test whether
documentation explains the structure we find. The cataloger is taken from the
two-letter prefix of the catalog name \citep{brezine2023}.

% ─────────────────────────────────────────────────────────────────────────────
\section{Unsupervised Clustering}
\label{sec:clustering}

We reduce the standardized feature matrix to two dimensions with
UMAP \citep{mcinnes2018} ($n\_neighbors=15$, $min\_dist=0.1$, Euclidean metric,
fixed seed) and cluster the embedding with HDBSCAN \citep{campello2013}
($min\_cluster\_size=10$, $min\_samples=5$). On the full corpus this yields three
clusters of 17, 159 and 443 khipus (10, 160 and 442 at $n=612$).

\paragraph{Silhouette space.}
Silhouette computed in the two-dimensional UMAP embedding is inflated by the
projection itself: on the full corpus it is 0.817, but only 0.062 when computed
in the 27-dimensional feature space where the clustering is meant to hold. We
report both, and base all null-model comparisons on feature space.

\subsection{Null models and sequential exclusions}
\label{sec:nulls}

We compare the observed silhouette with two null models, each simulated
$N=1{,}000$ times through the same UMAP + HDBSCAN pipeline: (i) a multivariate
Gaussian with the same mean and covariance as the standardized features, and
(ii) independent permutation of each feature, which preserves the (skewed,
zero-inflated) marginals but destroys joint structure. We report
$p=(1+\#\{\text{null}\ge\text{observed}\})/(N+1)$, whose smallest attainable value is $\approx0.001$.

\begin{table}[!htbp]
\centering
\caption{Clustering against null models ($N=1{,}000$ permutation and $1{,}000$
Gaussian simulations per subset). Silhouette and $p$ are given in the 27-D
feature space and, in parentheses, in the UMAP embedding.
$p\le0.001$ means no simulation reached the observed value.}
\label{tab:nulls}
\small
\begin{tabular}{lccccc}
\toprule
\textbf{Subset} & $n$ & $k$ & \textbf{Silhouette} & $p$ (perm.) & $p$ (Gauss.) \\
\midrule
All records & 619 & 3 & 0.062 (0.817) & 0.949 (0.159) & 0.036 ($\le0.001$) \\
Empty removed & 612 & 3 & 0.068 (0.793) & 0.823 (0.110) & 0.021 ($\le0.001$) \\
$<20\%$ unrecorded twist & 467 & 2 & 0.192 (0.591) & \textbf{0.009} (0.103) & 0.002 ($\le0.001$) \\
+ no contributor account & 430 & 3 & 0.107 (0.210) & 0.926 (0.913) & 0.043 (0.814) \\
\bottomrule
\end{tabular}
\end{table}

In feature space the permutation null is not rejected on the full corpus, after
removing empty records, or on the clean subset. The one rejection, at $n=467$,
is a two-cluster solution that does not reach significance in the UMAP space
($p=0.103$) and that vanishes when the 37 records of one contributor account are
removed. We therefore treat it as an isolated signal tied to those records
rather than as structure of the corpus. The Gaussian null is rejected weakly in
feature space in every subset ($p=0.002$--$0.043$); since the permutation null
preserves the zero-inflated marginals and is not rejected, we attribute that
departure to those marginals, not to joint structure. In the full and
empty-removed corpora the observed feature-space silhouette (0.062, 0.068) is
below the mean of the permutation null (0.259, 0.284).

\subsection{Clusters are twist-coding regimes}
\label{sec:colonial}

Cord-level twist coding separates the clusters (Table~\ref{tab:twist}): cluster~0
has no recorded S or Z cords at all (85.3\% blank or null, 14.7\% unspecified);
cluster~1 is mostly unspecified (68.1\%) with Z where twist is recorded; and
cluster~2 is predominantly S (90.1\%). The first version of this paper reported
85.3\% S-twist for a cluster; that figure is the \emph{blank} fraction.

\begin{table}[!htbp]
\centering
\caption{Cord-level twist coding by cluster (\%; full corpus, $n=619$ khipus).}
\label{tab:twist}
\begin{tabular}{lcrrrr}
\toprule
\textbf{Cluster} & \textbf{Khipus} & \textbf{S} & \textbf{Z} & \textbf{Unspecified} & \textbf{Blank/null} \\
\midrule
0 & 17 & 0.0 & 0.0 & 14.7 & 85.3 \\
1 & 159 & 1.5 & 27.6 & 68.1 & 2.8 \\
2 & 443 & 90.1 & 2.7 & 6.2 & 1.0 \\
\bottomrule
\end{tabular}
\end{table}

A depth-two decision tree on documentation indicators alone predicts cluster
membership with accuracy 0.914 against a majority-class baseline of 0.716 at
$n=619$, but only 0.716 against 0.745 at $n=430$
(\texttt{results/table\_b7\_doc\_tree.csv}). Predictability from missingness
disappears together with the clusters. We therefore read the three clusters as
three ways of recording, or not recording, twist, and not as culturally distinct
construction styles. That the dominant cluster is S-dominated may itself reflect
a recording convention (see Section~\ref{sec:limitations}). An earlier version
attributed one cluster to nineteenth-century European acquisition; museum
provenance and cataloger cannot be separated in this corpus, so we no longer
draw that inference.

\begin{figure}[!htbp]
    \centering
    \fig{\textwidth}{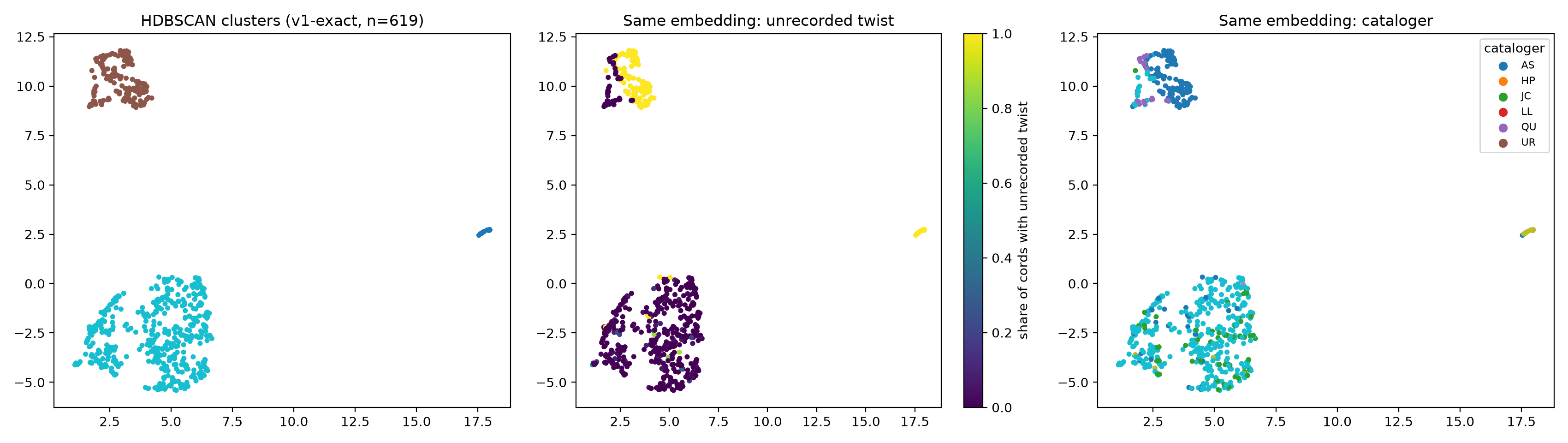}
    \caption{UMAP embedding of the khipus. \emph{Left:} HDBSCAN clusters.
    \emph{Middle:} the same embedding colored by the share of cords with
    unrecorded twist. \emph{Right:} colored by cataloger prefix. Cluster
    boundaries follow documentation completeness and cataloger.}
    \label{fig:clustering}
\end{figure}

\subsection{Sensitivity to the exclusion threshold}
\label{sec:thresh}

Table~\ref{tab:thresh} varies the unrecorded-twist threshold with the
contributor account removed. At the stricter thresholds (0.05 and 0.10) the
solution fragments into 8--10 clusters with feature-space silhouette near zero
and 20--27\% noise points; at 0.20 and 0.50 it returns to three clusters with
feature-space silhouette $\le0.125$. No threshold yields a well-separated
solution.

\begin{table}[!htbp]
\centering
\caption{Sensitivity to the unrecorded-twist exclusion threshold (records from
the single contributor account removed).}
\label{tab:thresh}
\begin{tabular}{cccccc}
\toprule
\textbf{Threshold} & $n$ & $k$ & \textbf{Sil.\ (UMAP)} & \textbf{Sil.\ (feat.)} & \textbf{Noise (\%)} \\
\midrule
0.05 & 395 & 8  & 0.415 & $-0.022$ & 19.7 \\
0.10 & 420 & 10 & 0.451 & 0.002    & 27.4 \\
0.20 & 430 & 3  & 0.210 & 0.107    & 3.3 \\
0.50 & 434 & 3  & 0.082 & 0.125    & 8.3 \\
\bottomrule
\end{tabular}
\end{table}

% ─────────────────────────────────────────────────────────────────────────────
\section{Provenance Classification}
\label{sec:classification}

\subsection{Setup}

We train a gradient-boosted tree classifier (XGBoost; 300 trees, depth 4,
learning rate 0.05) to predict region from the structural features, restricting
to the 120 khipus with an informative label (113 at $n=612$). Classes with fewer
than ten samples are grouped as ``Other''. With so few positives in some
classes, hyperparameter tuning is not meaningful, and we report no tuned model.
We evaluate with five-fold cross-validation, comparing stratified folds with
folds grouped by cataloger and by museum. Grouped schemes are not evaluable
for classes whose positives all belong to one group, and we report them as
such.

\subsection{Results}

Under stratified cross-validation the structural features give a weighted $F_1$
of 0.459 ($F_1=0.833$ for the Inka Late Horizon class); adding documentation
indicators raises it to 0.572 (0.944 for the imperial class). Both collapse
under grouped folds (weighted $F_1\le0.008$), and the imperial $F_1$ is not
evaluable there (Table~\ref{tab:cv}).

\begin{table}[!htbp]
\centering
\caption{Cross-validated weighted $F_1$ and Late Horizon $F_1$ by feature set
and fold scheme (XGBoost, 300 trees, depth 4). Grouped schemes are not
evaluable for the imperial class (---): all positives belong to one cataloger
(QU) and one museum (Dallas), so no positive is available for training when that
group is held out. Values differ slightly from Table~\ref{tab:imperial}, which
uses a smaller model (200 trees, depth 3).}
\label{tab:cv}
\small
\begin{tabular}{llcccc}
\toprule
 & & \multicolumn{2}{c}{$n=120$ (all records)} & \multicolumn{2}{c}{$n=113$ (empty removed)} \\
\textbf{Features} & \textbf{Folds} & $F_1^{w}$ & $F_1^{\text{imp}}$ & $F_1^{w}$ & $F_1^{\text{imp}}$ \\
\midrule
Structural & Stratified & 0.459 & 0.833 & 0.465 & 0.762 \\
 & By cataloger & 0.008 & --- & 0.007 & --- \\
 & By museum & 0.006 & --- & 0.007 & --- \\
\addlinespace
Structural + documentation & Stratified & 0.572 & 0.944 & 0.550 & 0.818 \\
 & By cataloger & 0.007 & --- & 0.007 & --- \\
 & By museum & 0.007 & --- & 0.007 & --- \\
\bottomrule
\end{tabular}
\end{table}

\subsection{The Late Horizon label tracks incompletely recorded Dallas khipus}
\label{sec:confound}

All 18 Late Horizon labels belong to the Dallas Museum of Art and to one
cataloger prefix (QU). Of these, 7 (KH0601--KH0607) have no cords, 7
(KH0593--KH0599) have cords but no knots or colors, and 4 (KH0589--KH0592)
carry color data (three of them with knots); 11 remain after removing empty
records and 3 in the clean subset ($n=430$).

Structural, twist-free and geometry-only feature sets give identical scores
(Table~\ref{tab:imperial}): in every case the classifier recovers exactly the 14
(7 at $n=612$) knotless or empty records and misses exactly the four with color
data, with two false positives at $n=120$ and none at $n=113$. Documentation
indicators do at least as well ($F_1=0.88$--$0.91$). The signal is incomplete
recording, not an imperial construction style. Because all positives share one
cataloger and one museum, grouped cross-validation cannot test generalization
for this class, and only three positives survive in the clean subset, so we do
not report imperial-class performance there.

\begin{table}[!htbp]
\centering
\caption{Recovering the Late Horizon label by feature set (five-fold CV,
XGBoost, 200 trees, depth 3). TP/FN/FP are counts; $p<0.01$ means that none of
100 label permutations reached the observed $F_1$ (not computed for the
eight-indicator set).}
\label{tab:imperial}
\small
\begin{tabular}{lcccccc}
\toprule
 & \multicolumn{3}{c}{$n=120$, $n_{\text{imp}}=18$} & \multicolumn{3}{c}{$n=113$, $n_{\text{imp}}=11$} \\
\textbf{Features} & TP/FN/FP & $F_1$ & $p$ & TP/FN/FP & $F_1$ & $p$ \\
\midrule
Structural (27) & 14/4/2 & 0.824 & $<0.01$ & 7/4/0 & 0.778 & $<0.01$ \\
Without twist/direction (21) & 14/4/2 & 0.824 & $<0.01$ & 7/4/0 & 0.778 & $<0.01$ \\
Geometry only (7) & 14/4/2 & 0.824 & $<0.01$ & 7/4/0 & 0.778 & $<0.01$ \\
Documentation (3) & 15/3/1 & 0.882 & $<0.01$ & 8/3/1 & 0.800 & $<0.01$ \\
Documentation (8) & 16/2/1 & 0.914 & --- & 9/2/0 & 0.900 & --- \\
\bottomrule
\end{tabular}
\end{table}

\subsection{SHAP attributions}

\begin{figure}[!htbp]
    \centering
    \fig{0.7\textwidth}{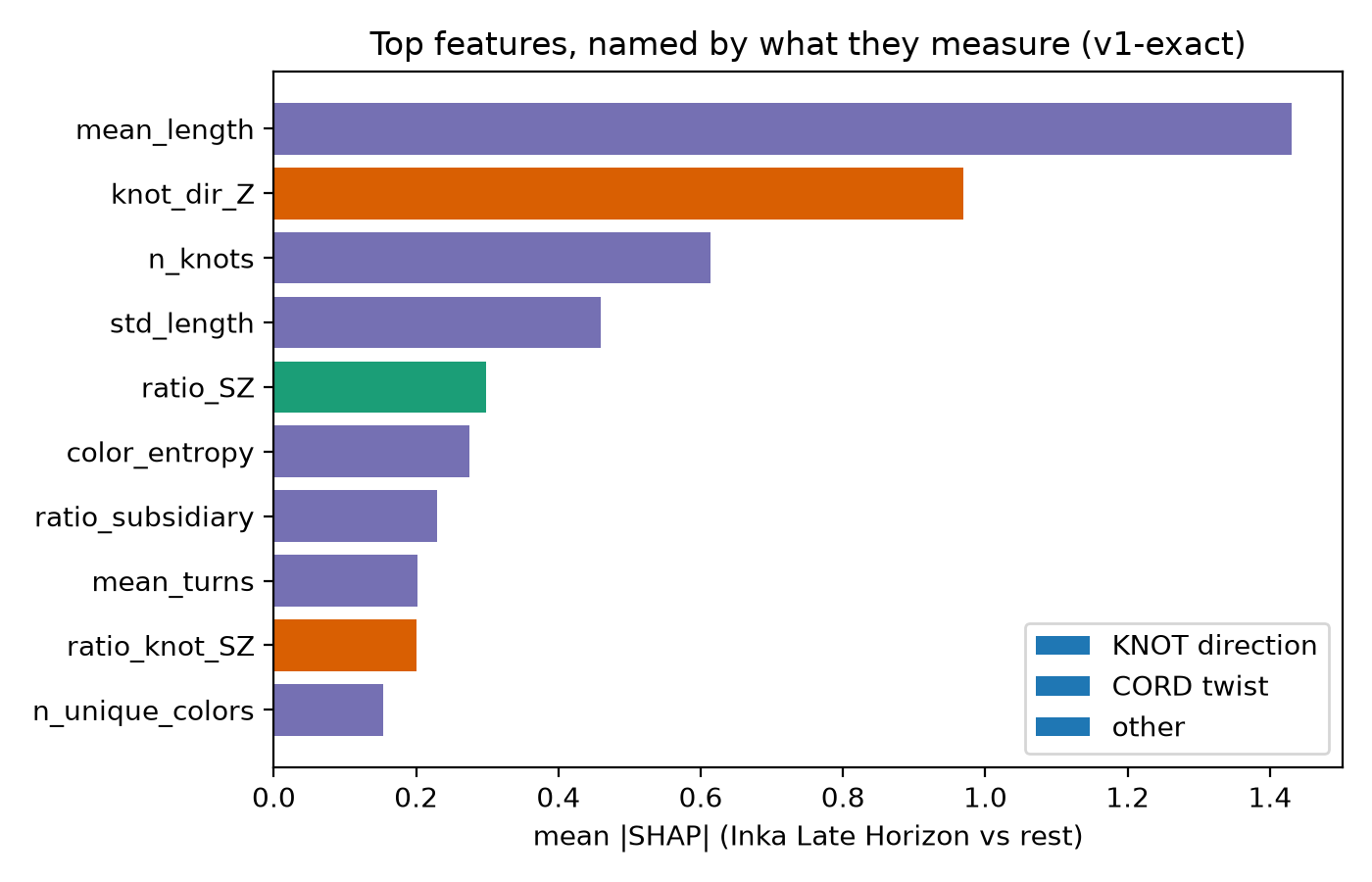}
    \caption{Mean absolute SHAP value of the top features for the Inka Late
    Horizon class, colored by what each feature measures.}
    \label{fig:shap}
\end{figure}

We compute SHAP attributions \citep{lundberg2017} for the Late Horizon model and
release them (Figure~\ref{fig:shap}), with each feature labeled by what it
measures (cord twist, knot direction, other); this corrects a mislabeling of
knot direction as cord twist in the first version. Given
Section~\ref{sec:confound}, we do not interpret the attributions as evidence of
manufacturing conventions: the model separates incomplete records from complete
ones.

\begin{figure}[!htbp]
    \centering
    \fig{0.7\textwidth}{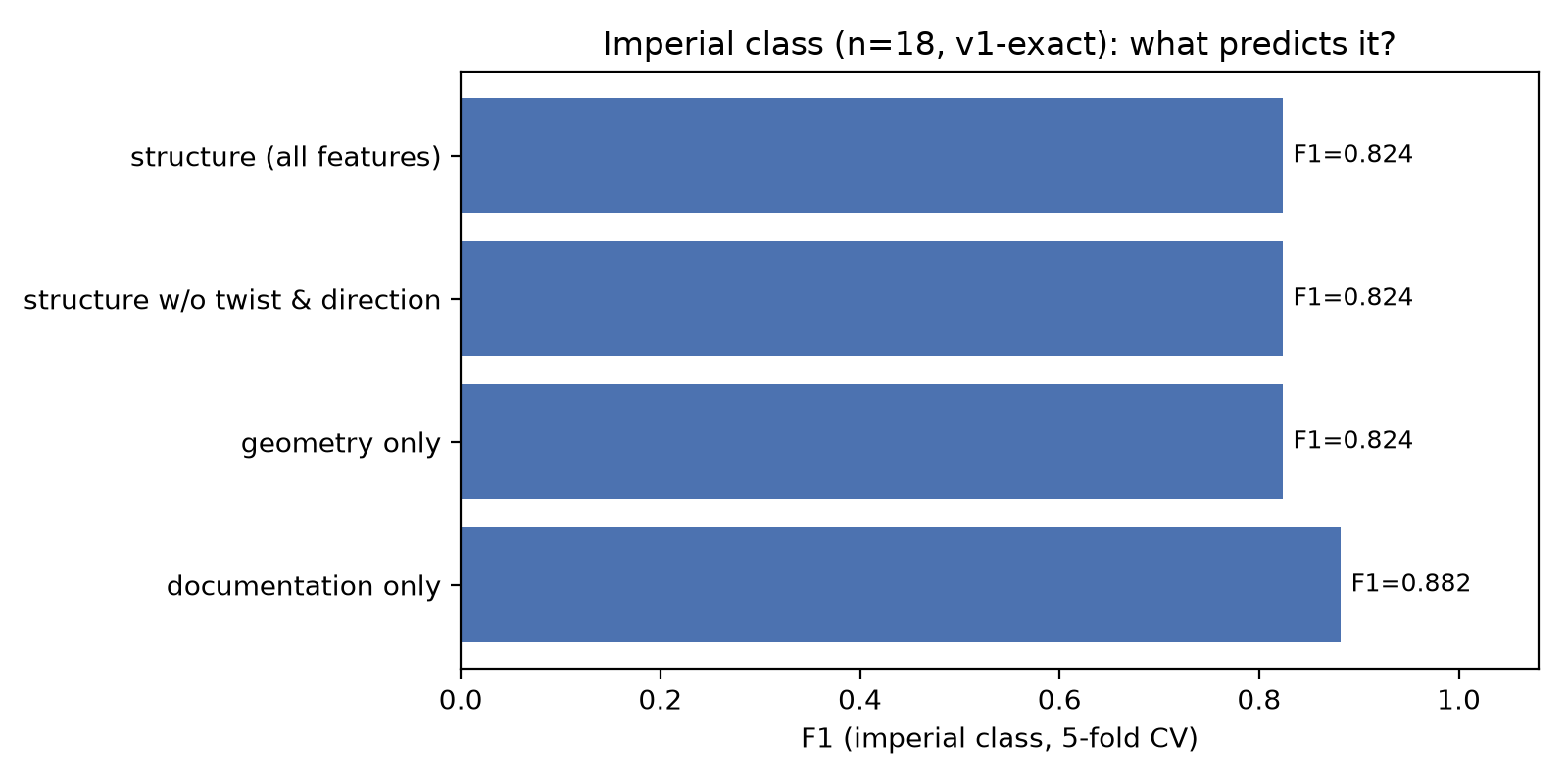}
    \caption{$F_1$ for the imperial class by feature set, with permutation $p$
    values.}
    \label{fig:imperial}
\end{figure}

% ─────────────────────────────────────────────────────────────────────────────
\section{Recount of the Santa Valley Recto/Verso Pattern}
\label{sec:santavalley}

\citet{medrano2018} matched six Santa Valley khipus (Harvard numbers
UR087--UR092; OKR numbers KH0323--KH0328; Museo Temple Radicati, Lima) to a
1670 colonial tribute census, arguing that cord attachment direction encodes
moiety affiliation. We ask whether the public OKR encodes the structural pattern
they describe.

Recto and verso are read directly from the OKR \texttt{cord.ATTACHMENT\_TYPE}
field, in which each cord is coded ``R'' (recto), ``V'' (verso), or ``U''
(unspecified), with no inference beyond this field. The six specimens are
selected by \texttt{OKR\_NUM BETWEEN 'KH0323' AND 'KH0328'}
(\texttt{results/table\_santa\_valley\_rv.csv}). Across all 1{,}078 cords we find
49.0\% recto and 51.0\% verso (excluding 249 unspecified cords), close to the
group-level ratio of about 47\%/53\% reported by \citet{medrano2018}; the two
ratios are computed over different cord sets and are not directly comparable.
Five of the six khipus are purely recto or purely verso, and a single specimen
(KH0326) is mixed (40 recto, 69 verso), matching the original identification of
one mixed khipu (Table~\ref{tab:santa}).

\begin{table}[!htbp]
    \centering
    \caption{Recto/verso cord counts for the six Santa Valley khipus,
    computed from the OKR. Five are pure; KH0326 is the sole mixed specimen.}
    \label{tab:santa}
    \begin{tabular}{lcc}
        \toprule
        \textbf{OKR No.} & \textbf{Recto} & \textbf{Verso} \\
        \midrule
        KH0323 & 310 & 0 \\
        KH0324 & 0 & 54 \\
        KH0325 & 0 & 207 \\
        KH0326 & 40 & 69 \\
        KH0327 & 0 & 93 \\
        KH0328 & 56 & 0 \\
        \bottomrule
    \end{tabular}
\end{table}

This recount uses the same transcriptions as the original study, so it is not an
independent verification: it shows that the public OKR encodes the pattern that
\citet{medrano2018} describe, not that new evidence confirms it. Nor do we
reproduce the numerical match to the 367-peso tribute total, which requires
positional knot-value decoding and is left for future work.

% ─────────────────────────────────────────────────────────────────────────────
\section{Limitations}
\label{sec:limitations}

Several limitations bound our conclusions. Only 120 of 619 khipus have an
informative region label, capping achievable performance and widening
cross-validation variance. The Late Horizon class ($n=18$; 11 after removing
empty records; 3 in the clean subset) is concentrated in one museum and one
cataloger prefix, so neither a within-group test nor grouped cross-validation is
possible, and classification performance for this class is not evidence of
cultural standardization. Zero-imputation of unrecorded twist, direction and
side fields makes the structural features partly proxies for documentation
completeness. The permutation null is not rejected in the full corpus or the
clean subset, so we claim no discrete structural typology; the single rejection
(at $n=467$) is not significant in UMAP space and depends on one contributor
account.

The unrecorded-twist threshold of 20\% is arbitrary (Table~\ref{tab:thresh}). The
S/Z contrast between records of one contributor account and other records from
the same two sites is unexplained: we treat those records as non-comparable
rather than recoding them, and a query to the OKR team is pending. That the
dominant cluster is 90\% S-twist may reflect a recording convention and not a
construction pattern, for the same reason. We also explored whether the
\emph{ordering} of knot types along cords carries provenance signal beyond
aggregate counts, encoding knot-type bigrams as TF--IDF features. They did not
improve classification: $\Delta F_1$ was $-0.009$ (stratified) and $0.000$
(cataloger-grouped) at $n=619$, and $-0.010$ and $+0.016$ at $n=612$, within
noise; a reconstructed 23-feature variant gives $-0.025$ and $-0.016$. This
suggests either that sequential order does not encode region or that the labeled
sample is too small to detect such a signal. Seven khipus (KH0593--KH0599) have
cords but no knots or colors because of a data-entry issue in the OKR and remain
in the analyses. None of these affects the Santa Valley recount, but they
substantially temper the clustering and provenance-classification results.

% ─────────────────────────────────────────────────────────────────────────────
\section{Ethical Considerations}
\label{sec:ethics}

Khipus are not merely archaeological data points; they are the intellectual
heritage of Andean peoples, and many specimens in the OKR were removed from
Peru during the colonial and post-colonial periods and now reside in foreign
institutions. The uneven documentation and cataloging we document in
Section~\ref{sec:colonial} reflect the history of how these objects were
collected, studied and recorded. We regard computational analysis as
a way to make Andean knowledge systems more legible and accessible---ideally to
descendant communities first---rather than as a means of extracting value from
cultural patrimony. We use the indigenous term \emph{khipu} throughout, rely
exclusively on an openly licensed community-maintained database, and release all
code and derived data openly so that researchers in Peru and across Latin
America can build on this work. We encourage future efforts to engage Quechua-
and Aymara-speaking communities and Peruvian institutions as collaborators, and
to support the eventual repatriation and community stewardship of these objects.

% ─────────────────────────────────────────────────────────────────────────────
\section{Conclusion}
\label{sec:conclusion}

We presented a reproducible audit of structural clustering in the Open Khipu
Repository and showed that its three clusters and the Late Horizon classifier
track recording completeness and record provenance rather than construction
style. In feature space, the cluster structure is not distinguishable from a
permutation null on the full corpus or on a clean subset, and the single
exception depends on records from one contributor account. The robust
contribution is methodological: corpus-wide structural claims about khipus must
be tested against null models, missingness and provenance controls before they
are interpreted. The recount of the Santa Valley recto/verso pattern is
unaffected, but, since it relies on the same transcriptions as the original
study, it is not independent evidence. We view structural pattern mining not as
a substitute for philological decipherment but as a complement: a way to
characterize the corpus, generate testable hypotheses, and quantify the biases
that any decipherment effort must confront. Future work includes the resolution
of the S/Z convention question with the OKR team, imperial-style khipus
documented by other catalogers and institutions, positional value decoding for
full numerical matching, sequence models over knot orderings, and the
integration of digitized colonial documents to extend the document-matching
approach beyond the single Santa Valley case.

% ─────────────────────────────────────────────────────────────────────────────
\section*{Data and Code Availability}

The complete pipeline is openly available:
\begin{itemize}
    \item Code and results: \url{https://github.com/mcontrerasmalpar-pixel/Khipu-ML}
    (notebook \texttt{02\_validations\_v2.ipynb}; tables in \texttt{results/},
    including the exclusion list and null-model tables)
    \item Notebook (first version): \url{https://kaggle.com/macmaky/khipu-ml}
    \item Dataset: Open Khipu Repository v2.1.0, DOI \texttt{10.5281/zenodo.18025748}
\end{itemize}

\section*{Acknowledgments}

The author thanks the Open Khipu Research Laboratory for maintaining the OKR
and for confirming the data-entry issue affecting KH0593--KH0599, the
maintainers of the original Harvard Khipu Database Project, and the Khipu Field
Guide project. The author also thanks Steven Fazzio
(\url{https://stevenfazzio.com/}) for a detailed public audit of the first
version's code and interpretation.

% ─── REFERENCES ──────────────────────────────────────────────────────────────

\end{document}